\documentstyle{article}
\newcommand{\ben}{\begin{enumerate}}
\newcommand{\een}{\end{enumerate}}
\newcommand{\bqt}{\begin{quote}}
\newcommand{\eqt}{\end{quote}}
\newcommand{\bc}{\begin{center}}
\newcommand{\ec}{\end{center}}
\newcommand{\bdes}{\begin{description}}
\newcommand{\edes}{\end{description}}
\newcommand{\ra}{$\rightarrow$}

\newcommand{\intn}{$i_{1},...,i_{n}$}

\newcommand{\bpack}{\begin{list}{$\bullet$}{\parsep 0pt \itemsep 4pt \topsep
4pt \parskip 0pt \partopsep 0pt \leftmargin 28pt}}
\newcommand{\epack}{\end{list}}
\newcommand{\bit}{\bpack}
\newcommand{\eit}{\epack}

\begin{document}
\title{Indefeasible Semantics and \\Defeasible Pragmatics\thanks{
I would like to thank David Beaver, Johan van Benthem, Paul Dekker,
Jan van Eijck, Jan Jaspars, Aravind Joshi, Alex Lascarides, Daniel Marcu,
Becky Passonneau, Henri\"{e}tte
de Swart, and Frank Veltman for helpful discussions and comments on
earlier versions of the paper.  The thoughtful comments by an
anonymous reviewer helped reshape the focus of the paper.  I also
profited from the comments from the seminar participants at the
University of Bielefeld and the University of Amsterdam.  I would
also like to thank those who responded to the pronoun interpretation
questionnaire whose results are discussed herein. Part of the work was
sponsored by project NF 102/62--356 (`Structural and Semantic
Parallels in Natural Languages and Programming Languages'), funded by
the Netherlands Organization for the Advancement of Research (N.W.O.).
}}

\author{Megumi Kameyama\\Artificial Intelligence Center and\\
The Center for the Study of Language and Information,\\SRI International\\
333 Ravenswood Ave., Menlo Park, CA 94025, U.S.A.}

\date{{\tiny 1995. To appear in Kanazawa, M., C. Pi\~{n}on, and H. de Swart,
eds.,\\ Quantifiers, Deduction, and Context. Stanford, CA: CSLI.}}
\maketitle

\section{Introduction}

An account of utterance interpretation in discourse needs to face the
issue of how the discourse context controls the space of interacting
preferences. Assuming a discourse processing architecture that
distinguishes the grammar and pragmatics subsystems in terms of
monotonic and nonmonotonic inferences, I will discuss how
independently motivated default preferences interact in the
interpretation of intersentential pronominal anaphora.

In the framework of a general discourse processing model that
integrates both the grammar and pragmatics subsystems, I will propose
a fine structure of the preferential interpretation in pragmatics in
terms of defeasible rule interactions. The pronoun interpretation
preferences that serve as the empirical ground draw from the survey
data specifically obtained for the present purpose.

\section{Discourse Processing Architecture}\index{discourse processing
architecture}

I will assume in this paper that a {\it discourse} is a sequence of
utterances produced (spoken or written) by one or more discourse
participants. {\it Utterances} are tokens of sentences or sentence
fragments with which the speakers communicate certain information, and
it is done in a {\it context}.  Utterance interpretation depends on
the context, and utterance meaning updates the context.

A specification of the complex interdependencies involved in utterance
interpretation is greatly facilitated if it is couched in a discourse
processing architecture that is both logically coherent and as closely
as possible an approximation of the human cognitive
architecture\index{human cognitive architecture} for discourse
processing. What are the major modules of the architecture, and what
types of inferences do they support?  I claim that the most
fundamental separation is between the spaces of {\it possibilities}
and {\it preferences}.

\subsection{Separating Combinatorics and Preferences}

There is an assumption in computational linguistics that combinatorics
should take precedence over preferences.  The wisdom is to maximize
the combinatoric space of utterance interpretation and to keep a firm
line between this space and the other, preferential, space of
interpretation.  Preferences are affected by computationally expensive
open-ended commonsense inferences.  Combinatorics determine all and
only possible interpretations, and preferences prioritize the
possibilities.\footnote{This separation of rule types does not imply a
sequential ordering between the two processing modules.  Different
rule types can be interleaved for interpreting or generating a
subsentential constituent.} Seen from another point of view,
combinatorics are {\it indefeasible} --- that is, never overridden by
commonsense plausibility, whereas preferences are {\it defeasible} ---
that is, can be overridden by commonsense plausibility. I will
henceforth assume that the grammar subsystem consists only of
indefeasible possibilities\index{indefeasible possibilities}, hence
monotonic\index{monotonicity}, whereas the pragmatics subsystem
consists mostly (or possibly entirely) of defeasible
preferences\index{defeasible preferences}, hence
nonmonotonic\index{nonmonotonicity}.\footnote{The same formal system
can be viewed from different viewpoints --- as a system of {\it
rules}, {\it constraints}, or {\it inferences}. Rules produce and
transform structures in a system, constraints reduce possible
structures, and inferences are used to reason about structures (e.g.,
manipulating assertions or drawing conclusions) as the ``logic'' in
the standard sense.  To take a prominent example, in the ``parsing as
deduction'' paradigm (Pereira and Warren, 1980), context-free rules
are also seen as deductive inference rules. The rule {\tt S \ra NP VP}
is translated into the inference rule {\tt NP(i,j) $\wedge$ VP(j,k)
$\rightarrow$ S(i,k)}. I will not adhere to one particular viewpoint
in this paper, and rather take advantage of the flexibility.}

An example of indefeasible rules of grammar in English is the
Subject-Verb-Object constituent order. The sentence {\it Coffee drinks
Sally} uttered in a normal intonation cannot mean ``Sally drinks
coffee'' despite the commonsense support.  An example of defeasible
preferences is the interpretation of the pronoun {\it he} in discourse
``{\it John hit Bill. He was severely injured.}'' The combinatoric
rule of pronoun interpretation would say that both John and Bill are
possible referents of {\it he}, while the preferential rule would say
that Bill is preferred here because it is more plausible that the one
who is hit gets injured rather than vice versa. Crucially, this
preference is overridden in certain contexts.  For instance, if Bill
is an indestructible cyborg, the preferred semantic value of {\it he}
would shift to John.

The inferential properties of the {\it grammar}
subsystem\index{grammar subsystem} as a space of possibilities are
well--illustrated in the so-called unification--based grammatical
formalisms\index{unification--based grammatical formalisms} (UBG). A
UBG system consists of context-free phrase structure constraints and
unification constraints. Maxwell and Kaplan (1993) describe how the
constraint interactions can be made efficient by exploiting the
following properties of a UBG system: (1) {\it monotonicity} --- no
deduction is ever retracted when new constraints are added, (2) {\it
independence} --- no new constraints can be deduced when two systems
are conjoined, (3) {\it conciseness} --- the size of the system is a
polynomial function of the input that it was derived from, and (4)
{\it order invariance} --- sets of constraints can be processed in any
order without changing the final result.\footnote{Grammar rules can be
seen from two viewpoints --- they {\it eliminate} as well as {\it
create} possibilities. The former applies when communication is seen
as incremental elimination of possible information states. The latter
applies when it is seen as incremental increase of information
content. I leave the choice open here.}

The inferential properties of the {\it pragmatics}
subsystem\index{pragmatics subsystem} are much less understood. Its
general features can be characterized as those of {\it preferential
reasoning}, a topic more studied in AI than in linguistics.  The
pragmatics subsystem contains sets of preference rules that, in
certain combinations, could lead to conflicting preferences. This
fundamental indeterminacy leads to the properties opposite from those
of the grammar subsystem: (1) {\it nonmonotonicity} --- preferences
can be canceled when overriding preferences are added, (2) {\it
dependence} --- new preferences may result when two pragmatic
subsystems are conjoined, (3) {\it explosion} --- the system size is
possibly an exponential (or worse) function of the input that it was
derived from, and (4) {\it order variance} --- changing the order in
which sets of preferences are processed may also change the final
result.  The key to a discourse processing architecture is to preserve
the above computational properties of the grammar subsystem while
striving for a maximal control of the preference interactions in the
pragmatics subsystem.\footnote{In contrast, the abduction--based
system\index{abduction--based system} (Hobbs et al., 1993) does not
separate grammar and pragmatics. All the rules are defeasible and
directly interact in one big module. (The defeasibility of grammar
rules is motivated by the fact of disfluencies in language use.) The
result is an increased computational complexity.}

Existing logical semantic theories employing dynamic interpretation
rules (e.g., Kamp, 1981; Heim, 1982; Groenendijk and Stokhof, 1991;
Kamp and Reyle, 1993) formalize the basic context dependence of
indefeasible semantics. While these theories predict the {\it
possible} dynamic interpretations of utterances, they are not
concerned with how to compute the relative preferences among them.
Lascarides and Asher (1993) extend the Discourse Representation
Theory\index{Discourse Representation Theory} (DRT) (Kamp, 1981) with
the interaction of defeasible rules for integrating a new utterance
content into the discourse information state. The input to their
defeasible reasoning is a fully interpreted DR
Structure\index{Discourse Representation Structure} (DRS), with all
the NPs already interpreted. The pragmatics subsystem I am concerned
with here also includes the defeasible rules for NP interpretation and
constituent attachments needed for DRS construction. The input to
pragmatics in the present proposal is a much less specified logical
form, and pragmatics kicks in {\it during} DRS construction.

\subsection{The Processing Architecture}\label{arch}

The discourse processing architecture\index{discourse processing
architecture} that I will assume in the background of the remainder of
this paper is this.\footnote{This architecture is in line with
Stalnaker's (1972:385) conception:
\bqt
The syntactical and semantical rules for a language determine an
interpreted sentence or clause; this, together with some features of
the context of use of the sentence or clause, determines a truth
value. An interpreted sentence, then, corresponds to a function from
contexts into propositions, and a proposition is a function from
possible worlds into truth values.
\eqt
}
\bit
\item Let {\em discourse} be a sequence of utterances,
$utt_1,\ldots,utt_n$.  We say that utterance $utt_i$ defines a {\em
transition relation} between the {\em input context} $C_{i-1}$ and the
{\em output context} $C_i$. Context $C$ is a multicomponent data
structure (see section \ref{context}).  The transition takes place as
follows:
\bit
\item Let {\it grammar} $G$ consist of rules of syntax and
semantics that assign each utterance $utt_i$ the {\it initial logical
form}\index{initial logical form} $\Phi_i$.

\item $\Phi_i$ represents a disjunctive set of underspecified
formulas\index{underspecified formulas}
containing unresolved references, unscoped quantifiers, and vague
relations. $\Phi_i$ is the weakest formula that packages a {\it
family} of formulas that covers the entire range of possible
interpretations of $utt_i$ (see section \ref{semantics}).

\item Let {\it pragmatics} $P$ consist of rules for specifying and
disambiguating $\Phi_i$ in context $C_{i-1}$. Ideally, $P$ outputs the
single {\it preferred interpretation}\index{preferred interpretation}
$\phi_i^k$ ($\phi_i^k$ is subsumed by $\Phi_i$ and there is no
$\phi_i^j$ that is preferred over $\phi_i^k$ and also subsumed by
$\Phi_i$), and integrating $\phi_i^k$ into context $C_{i-1}$ produces
the {\it preferred output context} $C_i$.  In a less felicitous case,
the rules of $P$ do not converge, resulting in multiple
interpretations and output contexts.
\eit
\eit

\subsection{Context}\label{context}\index{context}

My aim here is to introduce the basic components of the context $C$ in
the above discourse processing architecture that I assume in the
remainder of the paper.

Context $C_i$ is a 6-tuple $\langle\phi_i^k,D_i,A_i,I_i,L,K\rangle$
consisting of the fast-changing components,
$\langle\phi_i^k,D_i,A_i,I_i\rangle$, significantly affected by the
dynamic import of utterances and the slow-changing components,$\langle
L,K\rangle$, relatively stable in a given stretch of discourse
instance.  $\phi_i^k$ is the preferred interpretation (see section
\ref{arch}) of the last utterance $utt_i$ in a logical
form that preserves aspects of the syntactic structure of $utt_i$ ---
best thought of as a short-term register of the surface structure of
the previous utterance similar to the proposal by Sag and Hankamer
(1984).  $D_i$ is the {\it discourse model}\index{discourse model} ---
a set of information states that the discourse has been about, which
also incorporates the content of $\phi_i^k$.  $D_i$ contains sets of
situations, eventualities, entities, and relations among them,
associated with the evolving event, temporal, and discourse
structures.  $A_i$ is the {\it attentional state}\index{attentional
state} --- a partial order of the entities and propositions in $D_i$,
where the ordering is by {\it salience}\index{salience}.  $A_i$ is
separated from $D_i$ because the same $D_i$ may correspond to
different variants of $A_i$ depending on the particular sequence of
utterances in particular forms describing the same set of facts.
$I_i$ is the set of {\it indexical anchors}\index{indexical anchors}
--- the indexically accessible objects in the current discourse
situation --- for instance, the values of indexical expressions such
as {\it I, you, here,} and {\it now}. The slow-changing components are
the {\it linguistic knowledge}\index{linguistic knowledge} $L$ and
{\it world knowledge}\index{world knowledge} $K$ used by the discourse
participants. Although we know that discourse participants never share
exactly the same mental state representing these components of the
context, there must be a significant overlap in order for a discourse
to be mutually intelligible. For the purpose of this paper, I will
simply assume that context $C$ is sufficiently shared by the
participants.

The next section elaborates on the initial logical form $\Phi_i$ that
plays a crucial role of defining the grammar--pragmatics boundary in
the discourse processing architecture.

\section{Indefeasible Semantics}\label{semantics}\index{indefeasible semantics}

The initial logical form\index{initial logical form} (ILF) $\Phi$
represents the utterance's structure and meaning at the
grammar--pragmatics boundary\index{grammar--pragmatics boundary}.
This section discusses the general features of ILF with examples.

\subsection{General Considerations}

There are specific proposals for the ILF $\Phi$ in the computational
literature (e.g., Alshawi and van Eijck, 1989; Alshawi, 1992; Alshawi
and Crouch, 1992; Hwang and Schubert, 1992a, 1992b; Pereira and
Pollack, 1991).  Details in these proposals vary, but there is a
remarkable agreement on the general features.

The ILF $\Phi$ contains ``vague'' predicates and functions
representing {\it what} the utterance communicates.  Vague
predicates\index{vague predicates} and functions\index{vague
functions} represent various expression and construction types whose
interpretation depends on the discourse context. They include
unresolved referring expressions such as the pronoun {\it he},
unscoped quantifiers such as {\it each}, vague relations such as the
relation between the nouns in a noun--noun compound, unresolved
operators such as the tense operator {\it past} and the mood operator
{\it imperative}, and attachment ambiguities such as for
PP--attachments. The idea can also be extended to underspecify lexical
senses at the ILF level.  These predicates and functions generate
`assumptions' that need to be resolved or `discharged' in the union of
the discourse and sentence contexts. The ILF is thus {\it partial} and
{\it indefeasible} --- partial because it does not always have a truth
value, and indefeasible because further contextual interpretations
only prioritize possibilities and specify vagueness.

The ILF $\Phi$ also represents aspects of the utterance's surface
structure relevant to {\it how} the utterance communicates the
information content (e.g., the Topic--Focus
Articulation\index{Topic--Focus Articulation} of Sgall et al., 1986).
Such a syntax--semantics corepresentation could be achieved in either
of the two options: (1) the logical form is {\it structured},
representing aspects of phonological and surface syntactic structures
such as the grammatical functions of nominal expressions, linear
order, and topic--comment structure, or (2) the partial semantic
representation and the phonological and syntactic structures are
separately represented with mappings among corresponding parts.  In
this paper, the choice is arbitrary as long as certain syntactic
information is available at the logical form.

There is a general question of {\it how far} and {\it how soon} the
ILF gets specified and disambiguated by the pragmatics. The above
existing proposals in the computational literature assume that each
utterance is completely specified and disambiguated before the next
utterance comes in. This includes the integration of the utterance
content into the evolving discourse structure, event structure, and
temporal structure in the context, as discussed by Lascarides and
Asher (1993). An utterance's complete interpretation is not in general
available on the spot, however, and it often has to wait till some
more information is supplied in the subsequent discourse (Grosz et
al., 1986). It is also possible that only the information concerning
those entities that are significant or salient (or `in focus') in the
current discourse need to be fully specified and
disambiguated.\footnote{A comment by Paul Dekker.}  The present
discourse processing architecture allows such incremental and partial
specification and disambiguation of the information state along
discourse progression though this perspective is not explored in any
technical detail here.

In sum, the ILF represents the indefeasible semantics of an utterance
by leaving the following context--dependent interpretations
underdetermined: reference of nominal expressions, modifier
attachments, quantifier scoping, vague relations, and lexical
senses. The ILF also leaves open how the given utterance is integrated
into the temporal, event, and discourse structures in the context.

\subsection{Our Working Formalism}

I will use a simplified ILF in this paper. It is an underspecified
predicate logic in a davidsonian style --- a version of QLF (Kameyama,
1995) without the aterm--qterm distinction.  The ILF for the utterance
``{\it He made a robot spider}'' is as follows:
\bqt
$decl\ (past[\exists exy[make(e)\wedge Agent_{Subj}(e,x)\wedge
pro(x)\wedge he(x)\\\wedge Theme_{Obj}(e, y)\wedge indef\_sg(y)\wedge
spider(y)\\ \wedge nn\_relation(y,\lambda z(robot,z))]])$
\eqt
It contains the following vague predicates\index{vague predicates} and
functions\index{vague functions}:
\bit
\item unresolved unstressed pronoun ``he'' --- $pro(x)\wedge he(x)$
\item unscoped quantificational determiner ``a'' --- $indef\_sg(y)$
\item a vague relation for a noun-noun compound ``robot spider'' \\---
$spider(y)\wedge nn\_relation(y,\lambda z(robot,z))$ (a relation
between a spider entity and a robot property)
\item unresolved past tense --- $past(\psi)$
\item unresolved declarative mood --- $decl(\psi ')$
\eit
If the preferred interpretation of the utterance is that ``John'' made
a robot shaped like a spider, we have the following DRS--like logical
form:
\bqt
$\exists etxy[make(e)\wedge Time(e,t)\wedge Agent_{Subj}(e,x)\\\wedge
named(x,``john'')\wedge Theme_{Obj}(e,y)\wedge spider\_like(y)\wedge
robot(y)]$
\eqt
The interpretation is complete when the content is integrated into the
discourse, event, and temporal structures in the context.  These
structures are assumed to be in the discourse model $D$. The
pragmatics subsystem must make all of the preferential decisions
including NP interpretation and operator interpretation as well as
contextual integration.\footnote{I assume that various preferential
decisions are interleaved rather than sequentially ordered within
pragmatics.}

\subsection{Ambiguity and Underspecification}\index{ambiguity and
underspecification}

The initial logical form mixes both ambiguity and underspecification.
The choice is largely arbitrary when the number of possible
interpretations is exhaustively enumerable.  Whenever there are $n$
possible interpretations for a linguistic item or construction type,
we can have either (1) a disjunctive set of $n$ interpretations \intn,
 from which the pragmatics chooses the best, or (2) one underspecified
interpretation that the pragmatics further specifies.  Pragmatic
disambiguation and specification involve exactly the same kind of an
interplay of linguistic and commonsense preferences, and relative
preferences in disambiguation and specification are often
interdependent.

Consider {\it He made a robot spider with six legs}.  There is a
preference for the interpretation ``a robot spider with six legs''
over the alternative ``a male person with six legs''. This preference
is overridden in certain contexts --- for instance, if the person is a
fictional figure who can freely change the number of legs to be two,
four, or six, the alternative reading becomes equally plausible.  Note
that the attachment disambiguation and pronoun interpretation are
interdependent here.

When the number of possible interpretations cannot be exhaustively
enumerated, however, ambiguity and underspecification are not
interchangeable, and we must posit an {\it underspecified
relation}\index{underspecified relation} as a semantic primitive.  A
sufficient but not necessary condition for positing an underspecified
relation is this (Kameyama, 1995):\footnote{We have here an
operational criterion for separating out grammar and pragmatics.  It
leads to a discovery of cross--linguistic variation in the
grammar--pragmatics boundary.  Long--distance dependency is a case in
point (Kameyama, 1995).}
\bqt
An underspecified relation is posited when there is an open--ended set
of possible specific relations associated with a construction type,
and the interpretation is typically affected by {\it ad hoc} facts
known in the discourse context.
\eqt

A canonical example is the interpretation of noun--noun compounds such
as {\it elephant pen}. It could mean a pen shaped like an elephant, a
pen with elephant pictures on the body, a pen with a small toy
elephant glued on the top, or, depending on the context, a pen that
the speaker found on the ground when she was pretending to be an
elephant. All we can tell from the grammar of noun-noun compounds is
that it is a pen that has some salient relation with elephants. It
makes sense, then, to explicitly state in the grammar output the vague
notion of ``some salient relation'' as a primitive. This is the basic
motivation of the proposal for underspecified relations in the logical
form in the computational literature (e.g., Alshawi, 1990; Hobbs et
al., 1993).  The same thing goes with scope ambiguities. The number of
possible scopings is always bounded but possibly very large (on the
order of hundreds), and speakers are often unable to select a single
specific scoping, so the grammar should defer assigning specific
scopings to a sentence and give it to pragmatics (Hobbs, 1983; Reyle,
1993; Poesio, 1993).

In sum, with the ILF sealing off the space of grammatical reasoning,
the present discourse processing architecture magnifies the importance
of pragmatics in utterance interpretation. Pragmatics achieves
anaphora resolution, attachment disambiguation, quantifier scoping,
vague relation specification, and contextual integration all in one
module. Is there a system in the chaos? That is the question we turn
to now.

\section{Defeasible Pragmatics}\index{defeasible pragmatics}

This section discusses the features and examples of the defeasible
rules in the pragmatics subsystem.

\subsection{General Considerations}

By {\it defeasible}, I mean a conclusion that has to be retracted when
some additional facts are introduced.  This characterizes the {\it
preferential} aspect of utterance interpretation with the
nonmonotonicity property.  Grammatical reasoning\index{grammatical
reasoning} is governed by the Tarskian notion of valid inference in
standard logic --- ``Each model of the premises is also a model for
the conclusion.''  Pragmatic reasoning\index{pragmatic reasoning}
distinguishes among models as to their relevance or plausibility, and
is governed by the notion of plausible inference (Shoham, 1988) ---
``Each {\it most preferred} model of the premises is a model for the
conclusion.''  The preference can be stated in terms of default rules
as well, so the general reasoning takes the form of ``as long as no
exception is known, prefer the default.''  In utterance
interpretation, this form of reasoning chooses the best interpretation
 from among the set of possible ones.  The present focus is the
interpretation preferences of intersentential pronominal anaphora.

\subsection{Earlier Computational Approaches to Pronoun
Interpretation}\index{computational approaches to pronoun interpretation}

Computational research on pronoun interpretation has always recognized
the existence of powerful grammatical preferences, but there are
different views on their status in the overall processing
architecture.  Hobbs (1978) discussed the relative merit of purely
grammar--based and purely commonsense--based strategies for pronoun
interpretation. His grammar--based strategy that accounts for 98\% of
a large number of pronouns in naturally occurring texts simply could
not be extended to account for the remaining cases that only
commonsense reasoning can explain. He settled in a ``deeper'' method
that seeks a global {\it coherence} arguing that {\it coreference} can
be determined as a side--effect of coherence--seeking interpretation.
The abduction--based approach (Hobbs et al., 1993) is an example of
such a general inference system, where syntax--based preferences for
coreference resolution are used as the {\it last resort} when other
inferences do not converge.

Sidner's (1983) local focusing model used an {\it attentional}
representation level to mediate the grammar's {\it control} of
discourse inferences.\index{grammar's control of discourse inferences}
For each pronoun, there is an ordered list of potential referents
determined by local focusing rules, and the highest one that leads to
a consistent commonsense interpretation of the utterance is
chosen. Common sense has a veto power over grammar-based focusing in
the ultimate interpretation, but common sense {\it is} the last
resort, contrary to Hobbs's approach.  Carter (1987) implemented
Sidner's theory combined with Wilks's (1975) preferential semantics,
and reported the success rate of 93\% for resolving pronouns in a
variety of stories --- of which only 12\% relied on commonsense
inferences.

Grammar's role in the control of inferences was the original
motivation of the {\it centering model}\index{centering model} (Joshi
and Kuhn, 1979; Joshi and Weinstein, 1981).  The proposal was to use
the {\it monadic} tendency of discourse (i.e., tendency to be
centrally about one thing at a time) to control the {\it amount of
computation} required in discourse interpretation.  Grosz, Joshi, and
Weinstein (1983) proposed a refinement of Sidner's model in terms of
centering, and highlighted the crucial role of pronouns in linking an
utterance to the discourse context.  Subsequent work on centering
converged on an equally significant role of the main clause
SUBJECT\footnote{Grammatical functions will be in uppercase in order
to avoid the ambiguity of these words.} (Kameyama, 1985, 1986; Grosz,
Joshi, and Weinstein, 1986; Brennan, Friedman, and Pollard,
1987). Hudson D'Zurma (1988) experimentally verified that speakers had
a difficulty in interpreting a discourse where a centering prediction
was in conflict with commonsense plausibility, leading to a `garden
path' effect. An example from her experiment is: ``{\it Dick had a jam
session with Brad. He played trumpet while Brad played bass. ??He
plucked very quickly.}''
%The difficulty in this example is predicted from a clash
%between the centering preference to continue talking about Dick and
%the common sense that a trumpet cannot be plucked.
Centering models the local attentional state management in an overall
discourse model proposed by Grosz and Sidner (1986).

These computational approaches to discourse have recognized the
non--truth--conditional effects on utterance interpretation coming
 from the utterance's {\it surface structure} (i.e., phonological,
morphological, and syntactic structures). Although this aspect of
interpretation cannot be neglected in a discourse processing model,
its relevance to a logical model of discourse semantics and pragmatics
has remained unclear. It is worth pointing out that discourse
pragmatics\index{discourse pragmatics} in the above computational
approaches as well as in philosophy (e.g., Lewis, 1979; Stalnaker,
1980) has generally assumed a dynamic architecture. Would there be a
potential fit with the dynamic semantic theories in linguistics (e.g.,
Kamp, 1981; Heim, 1982; Groenendijk and Stokhof, 1991) in a way that
forms a basis for an integrated logical model of discourse semantics
and pragmatics?  In this paper, I propose a pronoun interpretation
model taking ideas from {\it both} computational and linguistic
traditions, and present it in such a way that it becomes tractable for
logical implementation.

\section{Pronoun Interpretation Preferences: Facts}\label{facts}

Pronoun interpretation must be carried out in an often vast space of
possibilities, somehow controlling the inferences with default
preferences coming from different aspects of the current context.
Pronouns such as {\it he, she, it} and {\it they} can refer to
entities talked about in the current discourse, present in the current
indexical context, or simply salient in the model of the world
implicitly shared by the discourse participants.  Since the problem
space is vast and complex, we need to narrow it down to come to grips
with interesting generalizations.  I will now limit our discussion to
the interpretation of the anaphoric use of {\it unstressed} male
singular third person pronoun {\it he} or {\it him} in English.

\subsection{Survey and the Results}

In 1993, I conducted a survey of pronoun interpretation preferences
using the discourse examples shown in Table 1.  These examples were
constructed to isolate the relevant dimensions of interest based on
previous work (see section \ref{discussion}).

One set of examples, A--H, involves pronouns that occur in the second
of two--sentence discourses. They were presented to competent (some
nonnative) speakers of English in the A-F-C-H-E-D-B-G order, avoiding
sequential effects of two adjancent similar examples. The speakers
were instructed to read them with no special stresses on words, and to
answer the who-did-what questions about pronouns in italics. The
answer ``unclear'' was also allowed, in which case, the speaker was
encouraged to state the reason. The total number of the speakers was
47, of which 10 were nonlinguist natural language researchers and 4
were nonnative but fluent English speakers.  The second set of
examples, I--L, are longer discourses. They were given to disjoint
sets of native English speakers, none of whom are linguists.

The examples fall under two general categories, as indicated in Table
1. One group isolates the {\it grammatical effects} by minimizing
commonsense biases. In these examples, it is conjectured that there is
no relevant commonsense knowledge that affects the pronoun
interpretation in question. The other group examines the {\it
commonsense effects} of a specific causal knowledge of hitting and
injuring in relation to the grammatical effects observed in the first
group.

\begin{footnotesize}
\begin{table}
\begin{tabular}{|ll|}\hline\hline
\multicolumn{2}{|l|}{Grammatical Effects:}\\\hline
A. & John hit Bill. Mary told {\it him} to go home. \\
B. & Bill was hit by John. Mary told {\it him} to go home. \\
C. & John hit Bill. Mary hit {\it him} too. \\
D. & John hit Bill. {\it He} doesn't like {\it him}. \\
E. & John hit Bill. {\it He} hit {\it him} back. \\
K. & Babar went to a bakery. He greeted the baker.\\
& {\it He} pointed at a blueberry pie. \\
L. & Babar went to a bakery. The baker greeted him.\\
& {\it He} pointed at a blueberry pie. \\\hline
\multicolumn{2}{|l|}{Commonsense Effects:}\\\hline
F. & John hit Bill. {\it He} was severely injured.\\
G. & John hit Arnold Schwarzenegger. {\it He} was severely injured.\\
H. & John hit the Terminator. {\it He} was severely injured.\\
I. & Tommy came into the classroom. He saw Billy at the door. \\
& He hit him on the chin. {\it He} was severely injured.\\
J. & Tommy came into the classroom. He saw a group of boys at the door.\\
& He hit one of them on the chin. {\it He} was severely injured.\\\hline\hline
\end{tabular}
\caption{Discourse Examples in the Survey}
\end{table}
\end{footnotesize}

Table 2 shows the survey results.  The $\chi^2_{df=1}$
significance\index{chi--square significance} for each example was
computed by adding an evenly divided number of the ``unclear'' answers
to each explicitly selected answer, reflecting the assumption that an
``unclear'' answer shows a genuine ambiguity. Preference is considered
{\it significant} if $p<.05$, {\it weakly significant} if $.05<p<.10$,
and {\it insignificant} if $.10<p$. Insignificant preference is
interpreted to mean ambiguity or incoherence. It follows from the
Gricean Maxim that ambiguity must be avoided in order for an utterance
to be pragmatically felicitous. An example with an insignificant
preference is thus infelicitous, and should not be generated.

It must be noted that the present survey results exhibit only one
aspect of preferential interpretation --- namely, the {\it final}
preference reached after an unlimited time to think. They do not
represent the {\it process} of interepretation --- for instance, a
number of speakers commented that they had to {\it retract} the first
obvious choice in example I. This garden--path effect verified in
Hudson D'Zurma's (1988) experiments does not show in the present
survey results.

\begin{footnotesize}
\begin{table}
\begin{tabular}{|l|lll|l|l|}\hline\hline
& \multicolumn{3}{|l|}{Answers} & $\chi^2_{df=1}$ & $p$\\\hline
A. & John 42 &  Bill 0 &  Unclear 5 & 37.53 & $p<.001$\\
B. & John 7 &  Bill 33 &  Unclear 7 & 14.38 & $p<.001$\\
C. & John 0 &  Bill 47 &  Unclear 0 & 47 & $p<.001$\\
D. & J. dislikes B. 42 &  B. dislikes J. 0 &  Unclear 5 & 37.53 & $p<.001$\\
E. & John hit Bill 2 &  Bill hit John 45 &  Unclear 0 & 39.34 & $p<.001$\\
K. & Babar 13 &  Baker 0 &  Unclear 0 & 13 & $p<.001$\\
L. & Babar 3 &  Baker 10 &  Unclear 0 & 3.77 & $.05<p<.10$\\\hline
F. & John 0 &  Bill 46 &  Unclear 1 & 45.02 & $p<.001$\\
G. & John 24 &  Arnold 13 &  Unclear 10 & 2.57 & $.10<p<.20$\\
H. & John 34 &  Terminator 6 &  Unclear 7 & 16.68 & $p<.001$\\
I. & Tommy 3 &  Billy 17 &  Unclear 1 & 9.33 & $.001<p<.01$\\
J. & Tommy 10 &  Boy 7 &  Unclear 3 & 0.45 & $.50<p<.70$\\\hline\hline
\end{tabular}
\caption{Survey Results}
\end{table}
\end{footnotesize}

\subsection{Discussion of the Results}\label{discussion}

The present set of examples highlights four major sources of
preference in pronoun interpretation --- {\it SUBJECT Antecedent
Preference, Pronominal Chain Preference, Grammatical Parallelism
Preference}, and {\it Commonsense Preference}. These are stated at a
descriptive level with no theoretical commitments. A theoretical
account of the same set of facts will be given in section
\ref{account}. Each source of preference is discussed below.

{\bf SUBJECT Antecedent Preference.}\index{subject antecedent
preference} A hierarchy of the preferred intersentential antecedent of
a pronoun has been proposed in the centering framework, which
basically says that the main clause SUBJECT is preferred over the
OBJECT (Kameyama, 1985,1986; Grosz et al., 1986).  This preference is
confirmed in examples A and B.\footnote{Some speakers indicated that
they had to assume additional facts in order to make a plausible
scenario --- for instance, in example A, ``Mary is a teacher, and she
sent John home as a punishment''. The speakers seem to want some more
information to make the judgment more conclusive.  What are the
relationships among these three people mentioned out of the blue? I
realize that impoverished examples of this sort rarely occur in our
real--life discourses. To sort out some rather delicate interplay of
preferences, however, we need to start out with simplified examples.
This is analogous to the use of the ``blocks world'' (i.e., the world
of blocks) in AI.}

The consistency of this preference across examples A and B
demonstrates that grammatical functions rather than thematic roles are
the adequate level of generalization. In both A and B, the thematic
roles of Bill and John in the first sentence are agent and theme (or
patient), respectively, but the switch in grammatical functions by
passivization causes the preferred interpretation to switch
accordingly.

Example C demonstrates the defeasibility of this preference in the
face of the parallelism induced by the adverb {\it too} as a side
effect of an indefeasible {\it conventional presupposition} (see
section \ref{account}).

{\bf Pronominal Chain Preference.}\index{pronominal chain preference}
This is the preference for a chain of pronouns across utterances to
corefer.\footnote{I will use the simple terminology of ``referent''
and ``coreference'' without committing to their realist connotation
because this does not affect the points I wish to make in this paper.}
Examples K and L are a minimal pair of structural effects without a
commonsense bias.  Their contrast shows the effect of grammatical
positions.  The SUBJECT--SUBJECT chain of pronouns (example K)
supports a significant coreference preference ($p<.001$), whereas the
OBJECT-SUBJECT chain (example L) supports a weakly significant {\it
noncoreference} preference ($.05<p<.10$) indicating a parallelism
effect below.

Example I shows that the causal knowledge also {\it in the end}
overrides a stretch of SUBJECT pronominal chain, but as noted above,
this example causes the speakers to first interpret the SUBJECT
pronouns to corefer, then {\it retract} the choice due to the
inconsistency with a causal knowledge. This processing tendency
indicates that the grammatical preference is processed faster than the
commonsense preference. We will come back to this issue later.

In example J, the strong preference for a SUBJECT pronominal chain is
undermined by the indefiniteness of the referent ({\it one of the
boys}) that the generic causal knowledge supports and by the
additional inference --- when one hits one of a group of boys, he
would be revenged by the group.  The grammar--based preference and
common sense are in a tie here, showing a genuine ambiguity
($.50<p<.70$).

{\bf Grammatical Parallelism Preference.}\index{grammatical
parallelism preference} There is a general preference for two adjacent
utterances to be grammatically parallel.  The parallelism requires,
roughly, that the SUBJECTs of two adjacent utterances corefer and that
the OBJECTs, if applicable, also corefer.  This preference is
demonstrated in example D that involves two pronouns.\footnote{Another
possible source of preference is the {\it causal link} between the two
described eventualities, John's hitting Bill ($e1$) and someone
disliking someone ($e2$).  The preferred interpretation supports the
causal link ``$e1$ {\it because} $e2$'', while the alternative
interpretation, which nobody took, supports ``$e1$ {\it therefore}
$e2$''.  These could be stated in terms of discourse relations of {\it
Explanation} and {\it Cause} (e.g., Lascarides and Asher, 1993). I'm
not aware of any empirical studies of this kind of preference
effects.} In example L, the parallelism preference overrides the
pronominal chain preference.

Example E shows the defeasibility of the parallelism preference in the
face of the presupposition triggered by adverb {\it back}.  An ``x hit
y back'' event conventionally presupposes that a ``y hit x'' event has
previously occurred, leading to the near-unanimous interpretation
``Bill hit John back.''\footnote{I suspect that the two speakers who
took the opposite interpretation used the sense of {\it back} close to
``again''.}

{\bf Commonsense Preference.}\index{commonsense preference} Examples
F--H illustrate the effect of a simple causal knowledge that dictates
the final interpretation over and above the grammatical preferences.
In example F, the SUBJECT Antecedent Preference is defeated by an
inference derived from the generic causal knowledge --- ``when X hits
Y, Y is normally hurt,'' and ``being injured is being hurt.'' Since
the example involves some ``normal'' fellows called John and Bill, it
applies with full force (46/47).

Examples G and H show what happens to this baseline default when the
described event involves some special individuals (fictitious or
nonfictitious) that the speakers have some knowledge about. In example
H, the preferred interpretation (34/47) swings to the one where the
normal fellow, John, is injured as a result of attempting to assault
the indestructible cyborg.\footnote{The Terminator is a cyborg played
by Arnold Schwarzenegger in a popular science--fiction movie.}  The
cyborg also could have been injured (6/47) (because the movie showed
that it {\it can} be destroyed after all).  In example G, John
attempts to assault a warm--blooded real person, Arnold, who seems a
little stronger than normal fellows.  Here, more speakers thought that
John was injured (24/47) than Arnold was (13/47), but this preference
is insignificant ($.10<p<.20$). It reflects the indeterminacy of
whether Arnold is a normal fellow or not, which affects the
applicability of the generic causal knowledge.\footnote{Of interest
here is the fact that the three speakers who knew {\it nothing} about
what a ``Terminator'' is {\it all} interpreted that John was injured
in example H. They clearly sensed ``something nasty and abnormal''
 from this name alone.}

\subsection{Descriptive Generalizations}

Table 3 summarizes the preference predicted by each of the four
sources discussed above and the final outcome verified in the
survey. We see the following general patterns of conflict
resolution:\index{conflict resolution patterns}
\ben
\item Conventional Presuppositions
(triggered by adverbs in examples C and E) and Commonsense Preferences
(examples F, G, and H) dictate the {\it final} preference.
\item Grammatical Preferences take charge in the {\it absence} of relevant
Commonsense Preferences (examples A--E, K, and L).
\item The SUBJECT Antecedent Preference overrides the Grammatical
Parallelism Preference when in conflict (see examples A and B), and
both are in turn stronger than the Pronominal Chain Preference
(example L).
\een

\begin{footnotesize}
\begin{table}
\begin{tabular}{|llll|l|l|}\hline\hline
& Subj.Pref. & Pron.Chain & Parallel. & Com.Sense & Outcome\\\hline
A. & John & --- & Bill & unclear & John\\
B. & Bill & --- & John & unclear & Bill\\
C. & John & --- & Bill & unclear & Bill$^{\clubsuit}$\\
D. & John--Bill? & --- & John--Bill & unclear & John--Bill\\
E. & John--Bill? & --- & John--Bill & unclear & Bill--John$^{\diamondsuit}$\\
K. & Babar & Babar & Babar & unclear & Babar\\
L. & Baker & Babar & Baker & unclear & Baker\\\hline
F. & John & --- & John & Bill & Bill\\
G. & John & --- & John & John/Arnold & John/Arnold\\
H. & John & --- & John & John & John\\
I. & Tommy & Tommy & Tommy & Billy & Billy$^{\spadesuit}$\\
J. & Tommy & Tommy & Tommy & Boy & Tommy/Boy\\\hline\hline
\end{tabular}

$\clubsuit$ --- due to the conventional presupposition triggered by adverb {\it
too}.\\
$\diamondsuit$ --- due to the conventional presupposition triggered by adverb
{\it back}.\\
$\spadesuit$ --- Tommy is the first choice, which is later retracted.
\caption{Preference Interactions: Facts}
\end{table}
\end{footnotesize}

The cases of indeterminate final preference\index{indeterminate
preference} in examples G and J are worth noting. This kind of an
indeterminate preference is infelicitous and uncooperative, which
should be avoided in discourse generation. The indeterminacy in
example G is due to the indeterminacy of Arnold being a normal person
subject to injury or an abnormally strong person who would not let
himself be injured. The indeterminacy in example J is due to the
conflict between the general causal knowledge about an injury caused
by hitting and the insalience of an indefinite referent as a possible
pronominal referent.

\section{Pronoun Interpretation Preferences: Account}\label{account}

Four major sources of preference have been identified in the above
pronoun interpretation examples. I propose that these sources
correspond to the data structures in the different context components
outlined in section \ref{context}.  The context components the most
relevant to the present discussion are the attentional state $A$, the
LF register $\phi$, and the discourse model $D$.

The main thrust of the present account is the general interaction of
preferences\index{preference interactions} that apply on different
context components. It explains the basic fact that preferences may or
may not be determinate. The present perspective of preference
interactions also extends and explains the role of the attentional
state in Grosz and Sidner's (1986) discourse theory.

\subsection{The Role of the Attentional State}

A discourse describes situations, eventualities, and entities,
together with the relations among them. The attentional
state\index{attentional state} $A$ represents a dynamically updated
snapshot of their {\it salience}\index{salience}. We thus assume the
property {\it salient} to be a primitive representing the {\it partial
order}\index{partial order} among a set of entities in $A$.\footnote{I
will not discuss the partial order of propositions.} The property {\it
salient} is gradient and relative.  A certain absolute degree of
salience may not be achieved by any entities in a given $A$, but there
is always a set of {\it maximally salient} entities, which is often,
but not necessarily, a singleton set.\footnote{Those entities that are
``inaccessible'' in the DRT sense do not participate in the salience
ordering, or even if they do, they are below a certain minimal
threshold of salience.} Thus it is crucial that a rule about the
single maximally salient entity in a given $A$ is only sometimes
determinate.

We will now recast some elements of the centering model in the present
discourse processing architecture.  In the input context $C_{i-1}$ for
utterance $utt_i$, the form and content ($\phi_{i-1}$) of the
immediately preceding utterance $utt_{i-1}$ occupy an especially
salient status.  The entities realized in $utt_{i-1}$ are among the
most salient subpart of $A_{i-1}$. I assume that this is achieved by a
general $A$-updating mechanism. One of the entities in $A_{i-1}$ may
be the $Center_{i-1}$, what the current discourse is {\it centrally
about}, hence the high salience:\footnote{In the centering model, the
entities realized in $\phi_{i-1}$ are the ``forward--looking centers''
(Cf), and $Center_{i-1}$ is the ``backward--looking center'' (Cb).}
\bdes
\item[CENTER] The Center is normally more salient than other entities in
the same attentional state.
\edes

At least two default linguistic hierarchies are relevant to the
dynamics of salience\index{salience dynamics}.\footnote{Consituents'
linear ordering and animacy are also relevant.} One is the {\it
grammatical function hierarchy}\index{grammatical function hierarchy}
(GF ORDER), and the other is the {\it nominal expression type
hierarchy}\index{nominal expression type hierarchy} (EXP ORDER).  The
GF ORDER in $utt_i$ predicts the relative salience of entities in the
{\it output} attentional state $A_i$ whereas the EXP ORDER in $utt_i$
predicts the relative salience of entities {\it assumed} in the {\it
input} attentional state $A_{i-1}$.\footnote{This order also
approximates the relative salience of entities in the {\it output}
attentional state, as demonstrated in part in example J.}  EXP ORDER
is also crucial to the management of the Center (EXP CENTER):
\bdes
\item[GF ORDER:] Given a hierarchy, [{\sc SUBJECT $>$ OBJECT $>$ OBJECT2 $>$
Others}], an entity realized by a higher ranked phrase is normally
more salient in the output attentional state.
\item[EXP ORDER:] Given a hierarchy, [{\sc Zero
Pronominal $>$ Pronoun $>$ Definite NP $>$ Indefinite
NP}],\footnote{There is a pragmatic difference between stressed and
unstressed pronouns, which should be accounted for by an independent
treatment of stress --- for example, in terms of a preference reversal
function (Kameyama, 1994b). This paper concerns only unstressed
pronouns.}  an entity realized by a higher--ranked expression type is
normally more salient in the input attentional state.
\item[EXP CENTER:] An expression of the highest ranked type
normally realizes the Center in the output attentional state.
\edes
EXP CENTER can be interpreted in two ways. One computes the
``highest--ranked type'' per utterance, sometimes allowing a
nonpronominal expression type to output the Center. The other takes it
to be fixed, namely, only the pronominals. The choice is empirical.
In this paper, I will take the second interpretation.

Since matrix subjects and objects cannot be omitted in
English,\footnote{Except in a telegraphic register.}  the
highest--ranked expression type is the (unstressed) pronoun (see
Kameyama, 1985:Ch.1).  From EXP ORDER, it follows that a pronoun {\it
normally} realizes a {\it maximally salient entity} in the input
attentional state.  A pronoun can also realize a submaximally salient
entity if this choice is supported by another overriding preference.
The grammatical features of pronouns also constrain the range of
possible referents --- for instance, a {\it he}--type entity is a male
agent. The maximal salience thus applies on the suitably restricted
subset of the domain for each type of pronoun.

The interactions of the above defeasible rules --- CENTER, GF ORDER,
EXP ORDER, and EXP CENTER --- account for various descriptive
generalizations. First, the SUBJECT Antecedent Preference follows from
GF ORDER and EXP ORDER --- SUBJECT is the highest ranked GF in the
first utterance, and a pronoun in the second utterance realizes the
maximally salient entity in the input $A$.  Second, the coreference
and noncoreference preferences in pronominal chains are accounted for.
The strong coreference preference for a SUBJECT--SUBJECT pronominal
chain (example K) comes from the fact that a SUBJECT Center is the
single maximally salient entity, which leads to a determinate
preference\index{determinate preference}. In contrast, an OBJECT
Center competes with the SUBJECT non--Center for the maximal salience,
which leads to an indeterminate preference\index{indeterminate
preference} based on salience alone (example L).  The indeterminacy is
resolved, to some extent, by the Grammatical Parallelism Preference
(section \ref{lf}).\footnote{This notion of the single maximally
salient entity corresponds to the ``preferred center'' Cp (Grosz et
al., 1986) that is determined solely by the GF ORDER. The difference
here is that it is determined by {\it both} the Center and GF ORDER,
predicting an indeterminacy in certain cases.}

The center transition types of ``establishing'' and ``chaining''
(Kameyama, 1985,1986) result from the interactions of CENTER, EXP
ORDER, and EXP CENTER.\footnote{What I have previously called {\it
retain} is now called {\it chain}. It covers both CONTINUE and RETAIN
technically distinguished by Grosz et al. (1986) and Brennan et
al. (1987).} The Center is ``established'' when a pronoun picks a
salient non--Center in the input context and makes it the Center in
the output context.  It is ``chained'' when a pronoun picks the Center
in the input context and makes it the Center in the output context.
Examples A--H are thus concerned with Center--establishing pronouns,
whereas examples I--L are concerned with Center--chaining
pronouns. These transition types are not the primitives that directly
drive preferences, however.

\subsection{The Role of the LF Register}\label{lf}

The grammatical parallelism of two adjacent utterances in discourse
affects the preferred interpretation of pronouns (Kameyama, 1986),
tense (Kameyama, Passonneau, and Poesio, 1993), and ellipsis (Pruest,
1992; Kehler, 1993). This general tendency warrants a separate
statement.  Parallelism is achieved, in the present account, by a
computation on the pair of logical forms, one in the LF register in
the context, and the other being interpreted.
\bdes
\item[PARA:] The LF register in the input context and the ILF being
interpreted seek maximal parallelism.\footnote{This statement is
intentionally left vague.  See Pruest's (1992) MSCD operation for a
general definition of parallelism preference, and my property--sharing
constraint (Kameyama, 1986) for a subcase relevant to pronoun
interpretation.}
\edes
The present perspective of rule interaction explains the
``property--sharing'' constraint on Center--chaining (Kameyama, 1986)
as follows.  GF ORDER, EXP ORDER, and PARA join forces to create a
strong grammatical preference for SUBJECT--SUBJECT coreference
(examples D,K). When they are in conflict, that is, when the maximally
salient entity is not in a parallel position, PARA is defeated
(examples A,B). When maximal salience is indeterminate, the
parallelism preference affects the choice (example L), leading to a
noncoreference preference for an OBJECT--SUBJECT pronominal chain.

\subsection{The Role of the Discourse Model}

The discourse model\index{discourse model} contains a set of
information states about situations, eventualities, entities, and the
relations among them. It also contains the evolving discourse
structure, temporal structure, and event structure. Both linguistic
semantics and commonsense preferences apply on the same discourse
model.

{\bf Lexically Triggered Presuppositions.} Adverbs {\it too} and {\it
back} trigger conventional presuppositions\index{ conventional
presuppositions}about the input discourse model. These presuppositions
are part of lexical semantics, thus indefeasible.

Adverb {\it too} triggers a presupposition that appears to seek
parallelism between an utterance in the context and the utterance
being interpreted. This is actually due to a general {\it similarity}
presupposition associated with {\it too}.  Consider each of the
following utterances immediately preceding ``John hit Bill too'':
``Mary hit Bill'', ``John hit Mary'', ``Mary kicked Bill'', ``John
kicked Mary'', ``Mary hit Jane'', and $?$``John called Bill''. What's
construed as `similar' in each case is a function of the particular
utterance pair, and intuitively, preferred pairs support more
similarities. Thus similarity comes in degrees, and a parallel
interpretation is due to the preference for a maximal similarity.

Adverb {\it back} triggers a presupposition for a {\it reverse}
parallelism. That is, the utterance ``Bill hit John back'' presupposes
that it occurred after ``John hit Bill''.

{\bf Commonsense Knowledge.} In contrast to the above rules that
belong to the linguistic knowledge, the commonsense
knowledge\index{commonsense knowledge} consists of all that an
ordinary speaker knows about the world and life. Formalizing common
sense is a major research goal of AI, where nonmonotonic reasoning has
been intensively studied. My goal here is not to propose a new
approach to commonsense reasoning but simply to highlight its
interaction with linguistic pragmatics\index{linguistic pragmatics} in
the overall pragmatics subsystem. We know one thing for sure --- there
will be a relatively small number of linguistic pragmatic rules that
systematically interact with an open--ended mass of commonsense
rules. Since the linguistic rules can be seen to {\it control}
commonsense inferences, our aim is to describe the former as fully as
possible, and specify how the ``control mechanism'' works. The
commonsense rules posited in connection to the examples in this paper
are thus meant to be exemplary. There will be different rules for each
new example and domain to be treated. The linguistic rules, however,
should be stable across examples and domains.

The single powerful causal knowledge at work in our examples is that
hitting may cause injury on the hittee but less likely on the hitter:
\bdes
\item[HIT:] When an agent x hits an agent y, y is normally hurt.
\edes
The effects of the Terminator and Arnold indicate that the
applicability of the HIT rule depends on the normality of the agents
involved. Relevant knowledge includes things like: An agent is
normally vulnerable, Arnold is a normal agent or an abnormally strong
agent, and Terminator is an abnormally strong agent.

\subsection{Account of the Rule Interactions}

We now state the preference interaction patterns\index{preference
interaction patterns} observed in Table 3 above. The SUBJECT
Antecedent Preference and Pronominal Chain Preference result from
CENTER, GF ORDER, EXP ORDER, and EXP CENTER.  These are the defeasible
{\it Attentional Rules} (ATT) stating the preferred attentional state
transitions. The Grammatical Parallelism Preference is PARA. This is
an example of the defeasible {\it LF Rules} (LF) stating the preferred
LF transitions. Conventional presuppositions triggered by {\it too}
and {\it back} are examples of the indefeasible {\it Semantic Rules}
(SEM) in the grammar constraining the interpretation in the discourse
model. The causal knowledge of hitting is HIT, with associated
knowledge ETC about agents, Terminator, and Arnold. These are examples
of the defeasible {\it Commonsense Rules} (WK) stating the preferred
discourse model. Table 4 identifies the rules that dominate the {\it
final} interpretation in examples A--L.

\begin{footnotesize}
\begin{table}
\begin{tabular}{|lllll|l|}\hline\hline
   & ATT  & LF   & WK      & SEM & Winner\\\hline
A. & John & Bill & unclear & --- & ATT\\
B. & Bill & John & unclear & --- & ATT\\
C. & John & Bill & unclear & Bill & SEM\\
D. & John--Bill? & John--Bill & unclear & --- & LF\\
E. & John--Bill? & John--Bill & unclear & Bill--John & SEM\\
K. & Babar & Babar & unclear & --- & ATT+LF\\
L. & Baker/Babar & Baker & unclear & --- & ATT+LF\\\hline
F. & John & John & Bill & --- & WK\\
G. & John & John & John/Arnold & --- & WK\\
H. & John & John & John & --- & WK\\
I. & Tommy & Tommy & Billy & --- & WK ({\tiny with difficulty})\\
J. & Tommy & Tommy & Boy(/Tommy) & --- & $??$\\\hline\hline
\end{tabular}

{\it Rules}: ATT=\{CENTER, GF ORDER, EXP ORDER, EXP CENTER\},
LF=\{PARA\}, WK=\{HIT, ETC\}, SEM=\{TOO, BACK\}.
\caption{Preference Interactions: Account}
\end{table}
\end{footnotesize}

{\bf General Features.} The first distinction among these rules is
defeasibility\index{defeasibility}. The SEM rules are indefeasible
whereas all other rules are defeasible. It is predicted that
indefeasible rules override all defeasible rules, as verified in
examples C and E.

What factor determines the interaction pattern among the defeasible
rules?  The three context components --- discourse model $D$,
attentional state $A$, and LF register $\phi$ --- all have their
preferred transitions. The $D$ preference results from {\it
proposition--level} (or ``sentence--level'')
inferences\index{proposition--level inferences} {\it directly}
determining the preferred model whereas the $A$ and LF preferences
result from {\it entity--level} (or ``term--level'')
inferences\index{entity--level inferences} only {\it indirectly}
determining the preferred model. We have seen that proposition--level
preferences, if applicable, generally override entity--level
preferences, albeit with a varying degree of difficulty.

Take two examples: (1) ``{\it John met Bill. He was injured.}'' and
(2) ``{\it John hit Bill. He was injured.}''  In (1), the ATT and LF
preference that the pronoun refers to John indirectly leads to the
preference that {\it John was injured}, which becomes the overall
preference in the absence of relevant WK rules.  In (2), relevant WK
rules directly support a proposition--level preference, {\it Bill was
injured}, which wins out (with a varying degree of difficulty). These
``flows of preference'' during an utterance interpretation are
illustrated below:
\ben
\item[(1)]
$[_S [_{NP}$ $he$]:{\small \{John$>$Bill\}} $was\ injured$]
$\Longrightarrow$ John was injured
\item[(2)]
$[_S [_{NP}$ $he$]:{\small \{John$>$Bill\}} $was\ injured$]:{\small \{Bill
was injured $>$ John was injured\}} $\Longrightarrow$ Bill was injured.
\een

{\bf Conflict Resolution Patterns.}\index{conflict resolution
patterns} We see a straightforward overriding pattern in examples A--H
involving ``Center--establishing'' pronouns: $ATT$ overrides $LF$, and
$WK$ overrides $ATT$ and $LF$.  Such an overriding relation can be
seen as a dynamic updating operation (;) (van Benthem et al., 1993)
--- preferences are evaluated in turn, the later ones overriding the
earlier ones: $LF;ATT;WK$.\footnote{$\phi ;\psi [X]$ means $\psi [\phi
[X]]$, where $p[X]$ means $X\cap [[p]]$ (update state $X$ with $p$).}
It may be the general pattern of ``changing preferences'' during
utterance interpretation.

Examples I--L involving ``Center--chaining'' pronouns show more or
less the same pattern except that the overriding gets more difficult
in some cases.  It is more difficult when a SUBJECT pronoun chain
supports a single maximally salient entity as in example I. This shows
that the LF and ATT preferences in fact join forces to interact with
the WK preferences.  This intuition is expressed with brackets:
$[LF;ATT];WK$.  The ``retraction'' observed in example I still fits
this pattern, but the increased difficulty in overriding is only
implicit.

Lascarides and Asher (1993) illustrate patterns of defeasible rule
interactions\index{defeasible rule interactions}. The two inference
patterns\index{inference patterns} most relevant here are the Nixon
Diamond\index{Nixon Diamond} and the Penguin Principle\index{Penguin
Principle} defined below ($\phi\rightarrow\psi$ means ``if $\phi$,
then indefeasibly $\psi$,'' and $\phi >\psi$ means ``if $\phi$, then
normally $\psi$.''):\footnote{In these definitions, I use the
notations from Asher and Morreau's (1993) Commonsense Entailment (CE)
logic as a theoretical meta--formalism without strictly adhering to
the CE ontology.}
\bdes
\item[Nixon Diamond] A conflict is unresolved resulting in an
ambiguity or incoherence:
$(\phi >\chi )\wedge(\psi >\neg\chi )\supset (\phi ,\psi >\chi\wedge\neg\chi)$.

\item[Penguin Principle] A conflict is resolved by the more specific principle
defeating the more general one:\footnote{It follows from Cautious
Monotonicity [A$\Rightarrow$B, A$\Rightarrow$C /
A,B$\Rightarrow$C]:\\$(\phi\rightarrow\psi )\wedge (\phi >\chi )\supset
(\phi ,\psi >\chi)$ because $(\phi\wedge\psi )\leftrightarrow\phi$.}
\\$(\phi\rightarrow\psi )\wedge (\phi >\chi )\wedge (\psi >\neg\chi
)\supset (\phi ,\psi >\chi)$.

\edes
On their account, any resolution of a conflict between two defeasible
rules should be a case of the Penguin Principle.  Does it explain all
the conflict resolution patterns observed in pronoun interpretation?

The Penguin Principle explains some of the conflict resolution
patterns --- for instance, the knowledge about specific agents,
Terminator and Arnold, override the generic causal knowledge about
hitting (examples G and H).  There may also be a remote conceptual
connection between the Penguin Principle and the pattern $[LF;ATT];WK$
in the following line --- grammatical preferences (ATT and LF) tend to
be more abstract than commonsense preferences (WK) about particular
types of eventualities, so the more specific support wins (Kameyama et
al., 1993). However, the LF, ATT, and WK rules apply on different data
structures, and cannot always be reduced to an indefeasible
implication ($\phi\rightarrow\psi$) as required in the Penguin
Principle. For instance, $hittee(x)$ can be $subject(x)$ or $\neg
subject(x)$ depending on the sentence structure, so we cannot say that
$hittee(x)$ implies $\neg subject(x)$ to derive the overriding pattern
in example F. What additional kinds of conflict resolution inferences
do we have then?

There are two additional conflict resolution patterns observed in the
present examples, which I will call the {\it Indefeasible
Override}\index{Indefeasible Override} and the {\it Defeasible
Override}\index{Defeasible Override}, defined below:
\bdes
\item[Indefeasible Override] An indefeasible principle overrides a
defeasible one: $(\phi\rightarrow\chi )\wedge (\psi >\neg\chi )\supset
(\phi ,\psi\rightarrow\chi )$.

\item[Defeasible Override] Given an explicit overriding relation, one
defeasible principle defeats another (even when $\psi >\neg\chi$):\\
 $(\psi ;\phi )\wedge (\phi >\chi )\supset (\phi ,\psi >\chi )$.

\edes
The Indefeasible Override follows from the monotonicity of classical
implication ($\phi\rightarrow\chi\supset\phi ,\psi\rightarrow\chi$),
and is an inherent principle in any nonmonotonic logic. It predicts
the fact that the SEM rules override all the defeasible rules
(examples C and E).  The Defeasible Override captures a certain {\it a
priori} given ``ranks'' or ``priorities'' among different sources of
information, using the {\it dynamic override} (;) operator, where
$\phi ;\psi$ means ``$\psi$ overrides $\phi$.''  It is motivated by
the view that preferences come from different sources, and are
associated with different ``degrees of defeasibility'' not necessarily
in terms of the Penguin Principle.\footnote{G\"{a}rdenfors and
Makinson's (1994) use of expectation ordering in preferential
reasonning achieves essentially the same effect.}  It enables us to
state the override pattern $[LF;ATT];WK$ while allowing a varying
degree of difficulty for WK's overriding. I hope to define a logical
system that axiomatizes these conflict resolution inferences.

\section{Further Questions}

A number of questions related to the present topic have not been
discussed. The first are {\it logical questions}.  What are the
connections with {\it update logics}\index{update logic} (e.g.,
Veltman, 1993)? We can see that the grammar subsystem supports {\it
straight updating}, whereas the pragmatics subsystem supports {\it
preferential updating} or {\it upgrading} (van Benthem et al., 1993).
The preference interaction patterns discussed here can perhaps be
formulated as fine--grained upgrading inferences during utterance
interpretation within the proposed utterance interpretation
architecture. Can my proposal be couched in a system of {\it
preferential dynamic logic}\index{preferential dynamic logic} that
combines elements of dynamic semantic theories and preferential models
(e.g., McCarthy, 1980; Shoham, 1988)?  Does the context as a
multicomponent data structure proposed here also support the general
contextual inferences such as {\em lifting} in the context
logic\index{context logic} (e.g., McCarthy, 1993; Buva\v{c} and Mason,
1993)?

There are also {\it computational questions}.
%The undecidability of predicate circumscription is well known (Davis, 1980).
Does the proposed discourse processing architecture with explicit
contextual control of inferences actually {\it help manage} the
computational complexity of the nonmonotonic reasoning in the
pragmatic rule interactions?

Finally, a {\it cognitive question} --- Does the proposed discourse
processing architecture naturally extend to a more elaborate
many--person discourse model that addresses the issue of coordinating
different {\it private} contexts (e.g., Perrault, 1990; Thomason,
1990; Jaspars, 1994)?

\section{Conclusions}

A discourse processing architecture with desirable computational
properties consists of a grammar subsystem representing the space of
possibilities and a pragmatics subsystem representing the space of
preferences. Underspecified logical forms proposed in the
computational literature define the grammar--pragmatics boundary.
Utterance interpretation induces a complex interaction of defeasible
rules in the pragmatics subsystem. Upon scrutiny of a set of examples
involving intersentential pronominal anaphora, I have identified
different groups of defeasible rules that determine the preferred
transitions of different components of the dynamic context.  There are
grammatical preferences inducing fast entity--level inferences only
indirectly suggesting the preferred discourse model, and commonsense
preferences inducing slow proposition--level inferences directly
determining the preferred discourse model. The attentional state in
the context supports the formulation of attentional rules that
significantly affect pronoun interpretation preferences. The observed
patterns of conflict resolution among interacting preferences are
predicted by a small set of inference patterns including the one that
assumes an explicitly given overriding relation between rules or rule
groups. In general, I hope that this paper has made clear some of the
{\it actual} complexities of interacting preferences in linguistic
pragmatics, and that the discussion has made them sufficiently sorted
out for further logical implementations.\footnote{In the longer
version of this paper (Kameyama, 1994a), a logical implementation of
the preferential rule interactions is proposed using prioritized
circumscription (McCarthy, 1980, 1986; Lifschitz, 1988), a nonmonotonic
reasoning formalism in AI.}

\section*{References}

\noindent
\begingroup
\parindent=-2em \advance\leftskip by2em
\parskip=5pt

Alshawi, Hiyan. 1990. Resolving Quasi Logical Forms.
{\it Computational Linguistics}, 16(3), 133--144.

Alshawi, Hiyan. ed. 1992. {\it The Core Language Engine}, The MIT
Press, Cambridge, MA.

Alshawi, Hiyan, and Richard Crouch. 1992. Monotonic Semantic
Interpretation.  In {\it Proceedings of the 30th Annual Meeting of the
Association for Computational Linguistics}, Newark, DE, 32--39.

Alshawi, Hiyan, and Jan van Eijck. 1989.  Logical Forms in the Core
Language Engine.  In {\it Proceedings of the 27th Annual Meeting of
the Association for Computational Linguistics}, Vancouver, Canada,
25--32.

Asher, Nicholas and Michael Morreau. 1993. Commonsense Entailment: A
Modal Theory of Nonmonotonic Reasoning. In {\it Proceedings of the
International Joint Conference on Artificial Intelligence}, Chambray,
France, 387--392.

van Benthem, Johan, Jan van Eijck, and Alla Frolova. 1993. Changing
Preferences. Report CS-R9310. Institute for Logic, Language and
Computation, University of Amsterdam.

Brennan, Susan, Lyn Friedman, and Carl Pollard. 1987.  A Centering
Approach to Pronouns.  In {\it Proceedings of the 25th Annual Meeting
of the Association for Computational Linguistics}, 155--162.

Buva\v{c}, Sa\v{s}a and Ian Mason. 1993. Propositional Logic of
Context. In {\it Proceedings of the 11th National Conference on
Artificial Intelligence}, 412--419.

Carter, David. 1987. {\it Interpreting Anaphors in Natural Language
Texts.} Ellis Horwood, Chichester, Sussex, UK.

%Davis, Martin. 1980. The Mathematics of Non--Monotonic Reasoning.
%{\it Artificial Intelligence}, 13(1,2), 73--80.

G\"{a}rdenfors, Peter and David Makinson. 1994. Nonmonotonic Inference
Based on Expectations. {\it Artificial Intelligence}, 65, 197--245.

Groenendijk, Jeroen and Martin Stokhof. 1991. Dynamic Predicate Logic. {\it
Linguistics and Philosophy}, 14, 39--100.

%Grosof, Benjamin N. 1991. Generalizing Prioritization. In Allen, J.,
%R. Fikes, and E. Sandwall, eds., {\it Principles of Knowledge
%Representation and Reasoning: Proceedings of the Second International
%Conference}, Morgan Kaufmann Publishers, San Mateo, CA, 289--300.

Grosz, Barbara, Aravind Joshi, and Scott Weinstein. 1983.  Providing a Unified
Account of Definite Noun Phrases in Discourse.  In {\it Proceedings of
the 21st Meeting of the Association of Computational Linguistics},
Cambridge, MA, 44--50.

Grosz, Barbara, Aravind Joshi, and Scott Weinstein. 1986.  Towards a
computational theory of discourse interpretation.  Unpublished
manuscript. [The final version to appear in {\it Computational
Linguistics} under the title ``Centering: A Framework for Modelling
the Local Coherence of Discourse'']

Grosz, Barbara and Candy Sidner. 1986. Attention, Intention, and the Structure
of Discourse. {\it Computational Linguistics}, 12(3), 175--204.

Heim, Irene. 1982. {\it The Semantics of Definite and Indefinite Noun
Phrases,} Ph.D. Thesis, University of Massachusetts, Amherst.

%Heim, Irene. 1983. On the Projection Problem for Presuppositions. In
%the {\it Proceedings of the West Coast Conference on Formal
%Linguistics.}

Hobbs, Jerry. 1978. Resolving Pronoun References. {\it Lingua}, 44,
311--338. Also in B.~Grosz, K.~Sparck-Jones, and B.~Webber, eds., {\it
Readings in Natural Language Processing}, Morgan Kaufmann, Los Altos,
CA, 1986, 339--352.

Hobbs, Jerry. 1983. An Improper Treatment of Quantification in
Ordinary English. In {\it Proceedings of the 21st Meeting of the
Association of Computational Linguistics}, Cambridge, MA, 57--63.

Hobbs, Jerry, Mark Stickel, Doug Appelt, and Paul Martin. 1993.
Interpretation as Abduction. {\it Artificial Intelligence}, 63, 69--142.

Hudson D'Zmura, Susan. 1988. {\it The Structure of Discourse and
Anaphor Resolution: The Discourse Center and the Roles of Nouns and
Pronouns.} Ph.D. Thesis, University of Rochester.

Hwang, Chung Hee, and Lenhart K. Schubert. 1992a. Episodic Logic: A
Comprehensive Semantic Representation and Knowledge Representation for
Language Understanding. Technical Report, Dept. of Computer Science,
University of Rochester, Rochester, NY.

Hwang, Chung Hee, and Lenhart K. Schubert. 1992b. Episodic Logic: A
Situational Logic for Natural Language Processing. In P.~Aczel,
D.~Israel, Y.~Katagiri, and S.~Peters, eds., {\it Situation Theory and
its Applications}, Volume 3, CSLI, Stanford, CA.

Jaspars, Jan. 1994. {\it Calculi for Constructive Communication,}
Ph.D. Thesis, University of Tilberg.

Joshi, Aravind, and Steve Kuhn. 1979. Centered Logic: The Role of Entity
Centered Sentence Representation in Natural Language Inferencing. In
{\it Proceedings of International Joint Conference on Artificial
Intelligence}, Tokyo, Japan, 435--439.

Joshi, Aravind, and Scott Weinstein. 1981. Control of Inference: Role of
Some Aspects of Discoruse Structure --- Centering. In {\it Proceedings
of International Joint Conference on Artificial Intelligence},
Vancouver, Canada, 385--387.

Kameyama, Megumi. 1985. {\it Zero Anaphora: The Case of Japanese.}
Ph.D. Thesis, Stanford University.

Kameyama, Megumi. 1986. A Property-sharing Constraints in Centering.
In {\it Proceedings of the 24th Annual Meeting of
the Association for Computational Linguistics}, New York, NY, 200--206.

Kameyama, Megumi. 1994a. Indefeasible Semantics and Defeasible
Pragmatics. CWI Report CS-R9441 and SRI Technical Note 544.

Kameyama, Megumi. 1994b. Stressed and Unstressed Pronouns:
Complementary Preferences. In P.~Bosch and R. van der Sandt,
eds., {\it Focus and Natural Language Processing}, Institute for Logic
and Linguistics, IBM, Heidelberg, 475-484.

Kameyama, Megumi. 1995.  The Syntax and Semantics of the Japanese
Language Engine. In R.~Mazuka and N.~Nagai, eds., {\it Japanese
Sentence Processing}, Lawrence Erlbaum Associates, Hillsdale, NJ,
153--176.

Kameyama, Megumi, Rebecca Passonneau, and Massimo Poesio.  Temporal
Centering. 1993. In {\it Proceedings of the 31st Meeting of the
Association of Computational Linguistics}, Columbus, OH, 70--77.

Kamp, Hans. 1981. A Theory of Truth and Semantic Representation. In
J.~Groenendijk, T.~Janssen, and M.~Stokhof, eds., {\it Formal Methods
in the Study of Language}, Mathematical Center, Amsterdam, 277--322.

Kamp, Hans, and Uwe Reyle. 1993. {\it From Discourse to Logic}, Kluwer
Academic Publishers, Dordrecht.

Kehler, Andrew. 1993. The Effect of Establishing Coherence in Ellipsis
and Anaphora Resolution.
In {\it Proceedings of the 31st Meeting of the
Association of Computational Linguistics}, Columbus, OH, 62--69.

Lascarides, Alex, and Nicholas Asher. 1993.  Temporal Interpretation,
Discourse Relations, and Commonsense Entailment. {\it Linguistics and
Philosophy}, 16, 437--493.

Lewis, David. 1979. Scorekeeping in a Language Game. {\it Journal of
Philosophical Logic}, 8, 339--359.

Lifschitz, Vladimir. 1988. Circumscriptive Theories: A Logic--Based
Framework for Knowledge Representation. {\it Journal of Philosophical
Logic}, 17(4), 391--442.

Maxwell, John, and Ronald Kaplan. 1993. The Interface between Phrasal
and Functional Constraints. {\it Computational Linguistics}, 19(4),
571--590.

McCarthy, John. 1980. Circumscription--A Form of Non-monotonic
Reasoning. {\it Artificial Intelligence}, 13(1,2), 27--39.

McCarthy, John. 1986. Applications of Circumscription to Formalizing
Commonsense Knowledge. {\it Artificial Intelligence}, 28, 89--116.

McCarthy, John. 1993. Notes on Formalizing Context. In {\it
Proceedings of the International Joint Conference on Artificial
Intelligence}, Chambray, France, 555--560.

Pereira, Fernando and Martha Pollack. 1991. Incremental
Interpretation. {\it Artificial Intelligence}, 50, 37--82.

Pereira, Fernando, and David Warren. 1980. Definite Clause Grammars
for Language Analysis. {\it Artificial Intelligence}, 13, 231--278.

Perrault, C. Ray. 1990. An Application of Default Logic to Speech Act
Theory. In P.~Cohen, J.~Morgan, and M.~Pollack, eds., {\it Intentions
in Communications}, MIT Press, Cambridge, MA, 161--185.

Poesio, Massimo. 1993. {\it Discourse Interpretation and the Scope of
Operators.} Ph.D. Thesis, University of Rochester.

Pruest, Hub. 1992. {\it On Discourse Structuring, VP Anaphora and Gapping.}
Ph.D. Thesis, University of Amsterdam.

Reyle, Uwe. 1993. Dealing with Ambiguities by Underspecification:
Construction, Representation, and Deduction. {\it Journal of
Semantics}, 10(2).

%de Rijke, Maaten. 1992. A System of Dynamic Modal Logic. To appear in
%the {\it Journal of Philosophical Logic}.

%de Rijke, Maaten. 1994. Meeting Some Neighbors. In van Eijck, Jan and
%Albert Visser, eds., {\it Logic and Information Flow}, The MIT Press,
%Cambridge, MA.

Sag, Ivan, and Jorge Hankamer. 1984. Toward a Theory of Anaphoric
Processing. {\it Linguistics and Philosophy}, 7, 325--345.

Sgall, Petr, Eva Haji\v{c}ov\'{a}, and Jarmila
Panevov\'{a}.. 1986. {\it The Meaning of the Sentence in its Semantic
and Pragmatics Aspects}, Reidel, Dordrecht and Academia, Prague.

Shoham, Yoav. 1988. {\it Reasoning about Change: Time and Causality
 from the Standpoint of Artificial Intelligence}, MIT Press, Cambridge,
MA.

Sidner, Candy. 1983.  Focusing in the comprehension of definite
anaphora.  In M.~Brady and R.~C. Berwick, eds., {\em Computational
Models of Discourse}, The MIT Press, Cambridge, MA, 267--330.

Stalnaker, Robert, C. 1972. Pragmatics. In Davidson and Harman, eds.,
{\it Semantics of Natural Language}, Reidel, Dordrecht, 380--397.

Stalnaker, Robert, C. 1980. Assertion. In P.~Cole, ed., {\it
Syntax and Semantics Vol.9: Pragmatics}, Academic Press, New York,
315--332.

Thomason, Richmond. 1990. Propagating Epistemic Coordination through
Mutual Defaults I. In R.~Parikh, ed., {\it Theoretical Aspects of
Reasoning about Knowledge, Proceedings of the Third Conference (TARK
1990)}, Morgan Kaufmann, Palo Alto, CA, 29--38.

Veltman, Frank. 1993. Defaults in Update Semantics. Manuscript,
Department of Philosophy, University of Amsterdam. To appear in the
{\it Journal of Philosophical Logic}.

Wilks, Yorick. 1975. A Preferential, Pattern--seeking Semantics for
Natural Language Inference. {\it Artificial Intelligence}, 6, 53--74.

\endgroup
\end{document}